\documentclass[10pt,twocolumn,letterpaper]{article}

\usepackage{wacv}
\usepackage{times}
\usepackage{epsfig}
\usepackage{graphicx}
\usepackage{amsmath}
\usepackage{amssymb}

\usepackage{subfig}
\usepackage{booktabs}
\usepackage{ragged2e}
\usepackage{siunitx}
\usepackage{multirow}
\usepackage{multicol}
\usepackage{url}

\usepackage{nw-notation}
\graphicspath{{images/}}

\newcommand{\eqname}{Equation}
\newcommand{\BScolor}{red}
\newcommand{\SBcolor}{blue}
\newcommand{\BS}[1]{\textcolor{\BScolor}{#1}}  % Best performing method.
\newcommand{\SB}[1]{\textcolor{\SBcolor}{#1}} % Second best, first runner up.

\makeatletter
\renewcommand{\paragraph}{%
  \@startsection{paragraph}{4}%
  {\z@}{2.75ex \@plus 1ex \@minus .2ex}{-1em}%
  {\normalfont\normalsize\bfseries}%
}
\makeatother

\wacvfinalcopy % *** Uncomment this line for the final submission

\def\wacvPaperID{502} % *** Enter the wacv Paper ID here

\ifwacvfinal\fi
\begin{document}

%%%%%%%%% TITLE
\title{Deep Cosine Metric Learning for Person Re-Identification}

% Authors at the same institution
%\author{First Author \hspace{2cm} Second Author \\
%Institution1\\
%{\tt\small firstauthor@i1.org}
%}
% Authors at different institutions
\author{Nicolai Wojke \\
German Aerospace Center (DLR)\\
{\tt\small nicolai.wojke@dlr.de}
\and
Alex Bewley \\
University of Oxford\\
{\tt\small bewley@robots.ox.ac.uk}
%Queensland University of Technology\\
%{\tt\small alex.bewley@hdr.qut.edu.au}
}

\maketitle
\ifwacvfinal\thispagestyle{empty}\fi

\begin{abstract}
Metric learning aims to construct an embedding where two extracted features
corresponding to the same identity are likely to be closer than features from different identities.
This paper presents a method for learning such a feature space where the cosine similarity is effectively optimized through a simple re-parametrization of the conventional softmax classification regime.
At test time, the final classification layer can be stripped from the network to facilitate nearest neighbor queries on unseen individuals using the cosine similarity metric.
This approach presents a simple alternative to direct metric learning
objectives such as siamese networks that have required sophisticated
pair or triplet sampling strategies in the past.
The method is evaluated on two large-scale pedestrian re-identification
datasets where competitive results are achieved overall.
In particular, we achieve better generalization on the test set compared to
a network trained with triplet loss.

%We present a re-parametrization of the softmax classifier that produces
%compact clusters in representation space 
%We propose a novel approach to learn such a space where the angle between
%samples strongly reflects semantic similarity.
%Our key insight is based on a relationship between cosine similarity
%and von Mises-Fisher distribution which allows us to place metric learning
%into a classification framework.
%\textcolor{red}{TODO: Abstract needs a revision. The rest is a copy from our previous paper}
%Our approach offers several benefits over existing deep metric learning methods:
%(1) Our approach does not require mining of informative data samples,
%making it easier to implement and scalable to large datasets.
%(2) We can apply a well-established, easy-to-optimize classification loss,
%allowing for effective stochastic gradient descent optimization
%without losing interpretability of the representation space.
%We demonstrate the effectiveness of our method for the task of person
%re-identification where we surpass the state of the art on various large scale
%benchmarks.

\end{abstract}

\newcommand{\andothers}{\etal}
\newcommand{\figurespacing}{\vspace{-0.25cm}}
\newcommand{\figurespacingSmall}{\vspace{-0.2cm}}

\section{Introduction}
\label{sec:metric-learning-intro}

Person re-identification is a common task in video surveillance where a given query
image is used to search a large gallery of images potentially containing the same person.
As gallery images are usually taken from different cameras at
different points in time, the system must deal with pose variations, different
lighting conditions, and changing background.
Furthermore, direct identity classification is prohibited in this scenario because individuals in the gallery collected at test time are not contained in the training set.
Instead, the re-identification problem is usually addressed within a metric learning framework. Here the goal is to learn a feature representation -- from a set of separate training identities -- suitable for performing nearest neighbor queries on images and identities provided at test time. Ideally, the learnt feature representation should be invariant to the aforementioned nuisance conditions while at the same time follow a predefined metric where feature similarity corresponds to person identity.

%Instead, the re-identification problem is addressed in a metric learning framework. On a set of separate training identities, a feature representation is learned to reflect the task objective. At test time, the search for images of the same identity is performed using nearest neighbor queries. Ideally, the learnt feature representation should be invariant to the aforementioned nuisance conditions while at the same time follow a predefined metric corresponding to person identity suitable for nearest neighbor queries.

Due to the annotation effort that is necessary to set up a person
re-identification dataset, until recently only a limited amount of labeled
images were available.
This has changed with publication of the Market
1501~\cite{zheng2015scalable} and MARS~\cite{zheng2016mars} datasets.
MARS contains over one million images that have been annotated in a
semi-supervised fashion.
The data has been generated using a multi-target tracker that extracts short,
reliable trajectory fragments that were subsequently annotated to consistent
object trajectories.
This annotation procedure not only leads to larger amount of data, but also
puts the dataset closer to real-world applications where people
are more likely extracted by application of a person detector rather than
manual cropping.

\begin{figure}[t]
	\centering
	\includegraphics[width=\linewidth]{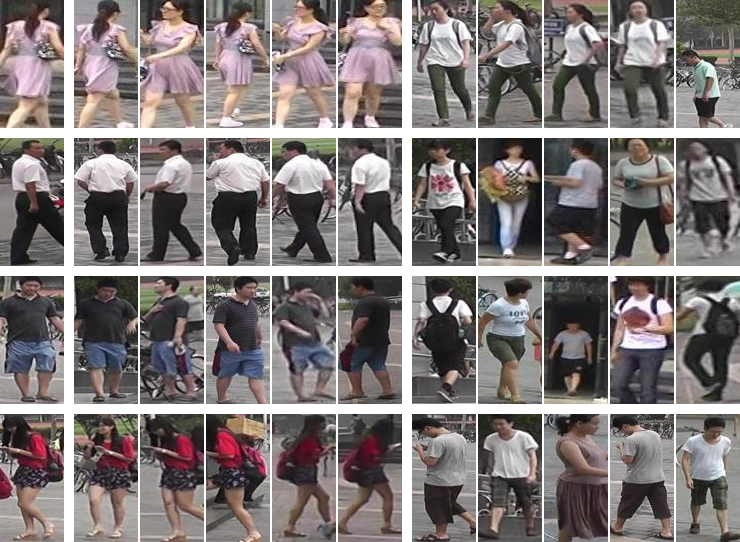}
	\caption[Example queries from Market~1501 test set]{%
        The proposed classifier successfully learns a metric representation
        space that is robust to articulation, lighting, and background
        variation.
        For each query image the five most similar and dissimilar images
        are shown.
        %The first image in each row shows a query image. The second block
        %shows the five most similar images in the gallery. The five most
        %dissimilar images are shown in the final block.
		%Example of successful queries generated from
		%Market~1501~\cite{zheng2015scalable} test gallery.
		%The first image in each row shows the query image. The second block
		%shows the five most similar images in the gallery.
		%The third block shows the five most dissimilar images in the gallery.
		%\textcolor{red}{TODO: Caption should be more like a mini abstract by briefly mentioning the metric property.}
	}
	\label{fig:market-query-good}

\figurespacing{}
\end{figure}

Much like in other vision tasks, deep learning has become the predominant
paradigm to person re-identification since the advent of larger datasets.
Yet, the problem remains challenging and far from solved.
In particular, there is an ongoing discourse over the performance of direct
metric learning objectives compared to approaching the training procedure
indirectly in a classification framework.
%The difference between the two methodologies is as follows.
Whereas metric learning objectives encode the similarity metric directly into
the training objective, classification-based methods train a classifier on the
set of identities in the training set and then use the underlying feature
representation of the network to perform nearest neighbor queries at test time.
On the one hand, in the past direct metric learning objectives have suffered
from undesirable properties that can hinder optimization, such as
non-smoothness or missing contextual information about the neighborhood
structure~\cite{rippel2015metric}.
On the other hand, these problems have been approached with success in
more recent publications~\cite{oh2016deep,hermans2017defense}. %ALEX: I feel the remainer doesn't really follow a logical order/argument
Nevertheless, with similarity defined solely based on class membership,
it remains arguable if direct metric learning has a clear advantage over
training in a classification regime.
In this setting, metric learning is often reduced to minimizing the distance
between samples of the same class and forcing a margin between samples of
different classes~\cite{chopra2005learning,hermans2017defense}.
A classifier that is set up with care might decrease intra-class variance and
increase inter-class variance in a similar way to direct metric learning
objectives.
%If the classifier is set up to follow a predefined metric on the representation
%space \textit{directly}, this classifier likely decreases intra-class
%variation and increase inter-class variation in the same way.

Inspired by this discussion, the main contribution of this paper is the
unification of metric learning and classification. More specifically, we
present a careful but simple re-parametrization of the softmax classifier that
encodes the metric learning objective directly into the classification task.
Finally, we demonstrate how our proposed cosine softmax training extends the
effectiveness of the learnt embedding to unseen identities at test time within
the context of person re-identification.
Source code of this method is provided in a GitHub
repository\footnote{\url{github.com/nwojke/cosine_metric_learning}}.

\section{Related Work}
\label{sec:metric-learning-literature}

%The following literature review focuses on metric learning and person
%re-identification, restricted to methods in a deep learning context.

\paragraph{Metric Learning}

Convolutional neural networks (CNNs) have shown impressive performance on large
scale computer vision problems and the representation space underlying these
models can be successfully transferred to tasks that are different from the
original training objective~\cite{donahue2014decaf, sharif2014cnn}.
Therefore, in classification applications with few training examples a
task-specific classifier is often trained on top of a general purpose feature
representation that was learned beforehand on
ImageNet~\cite{krizhevsky2012imagenet} or MS COCO~\cite{lin2014microsoft}.
There is no guarantee that the representation of a network which has been
trained with a softmax classifier can directly be used in an image retrieval
task such as person re-identification,
because the representation does not necessarily follow a certain (known) metric
to be used for nearest-neighbor queries.
Nevertheless, several successful applications in face verification and person
re-identification exist~\cite{%
    taigman2014deepface, xiao2016learning, zheng2016person}.
In this case, a softmax classifier is trained to discriminate the identities
in the training set. When training is finished, the classifier is stripped of
the network and distance queries are made using cosine similarity or Euclidean
distance on the final layer of the network.
If, however, the feature representation cannot be used directly, an alternative
is to find a metric subspace in a post processing
step~\cite{koestinger2012large,liao2015person}.
%KISSME:
%    Learns mahalanobis metric.
%    Objective is to find a linear subspace where the distance between
%    same and different identities is small and large.
%    Proposed for triplet loss, large margin nearest neighbor by Weinberger
%XQDA:
%    Extension of KISSME to multi-view problems. Cross-view quadratic
%    discriminant analysis, takes into account the variance in the projected
%    space.

Deep metric learning approaches encode notion of similarity directly into the
training objective.
The most prominent formulations are siamese networks with
contrastive~\cite{chopra2005learning} and
triplet~\cite{weinberger2009distance} loss.
The contrastive loss minimizes the distance between samples of the same class
and forces a margin between samples of different classes.
Effectively, this loss pushes all samples of the same class towards
a single point in representation space and penalizes overlap between different
classes.
The triplet loss relaxes the contrastive formulation to allow
samples to move more freely as long as the margin is kept.
Given an anchor point, a point of the same class, and a point of a
different class, the triplet loss forces the distance to the point of
the same class to be smaller than the distance to the point of the different
class plus a margin.
%The triplet loss forces the distance of examples of the same class
%to be smaller than the distance to examples of different classes.
%Therefore, the triplet loss relaxes the contrastive formulation by allowing
%samples to move freely within a specified margin.

Both the contrastive and triplet losses have been applied successfully to metric learning
problems~(e.g.,~\cite{schroff2015facenet, varior2016gated, hermans2017defense}),
but the success has long been dependent on an intelligent pair/triplet sampling
strategy.
Many of the possible choices of pairs and triplets that one can generate from
a given dataset contain little information about the relevant structures by
which identities can be discriminated.
%task objective.
If the wrong amount of hard to distinguish pairs/triplets are incorporated
into each batch, the optimizer either fails to learn anything meaningful or
does not converge at all.
Development of an effective sampling strategy can be a complex and
time consuming task, thus limiting the practical applicability of
siamese networks.

A second issue related to the contrastive and triplet loss stems from the
hard margin that is enforced between samples of different classes. The
hard margin leads to a non-smooth objective function that is harder to
optimize, because only few examples are presented to the optimizer
at each iteration and there can be strong disagreement % in the task objective
between different batches~\cite{rippel2015metric}.
These problems have been addressed recently.
%Qian~\andothers~\cite{qian2015fine} propose to mitigate this problem by solving the
%triplet loss formulation in multiple stages, where each stage operates on a
%smaller number of highly informative examples.
For example, Song~\andothers~\cite{oh2016deep} formulate a smooth upper bound
of the original triplet loss formulation that can be implemented by drawing
informative samples from each batch directly on a GPU.
A similar formulation of the triplet loss where the hard margin is replaced
by a soft margin has shown to perform well on a person re-identification
problem~\cite{hermans2017defense}.

Apart from siamese network formulations, the magnet loss~\cite{rippel2015metric}
has been formulated as an alternative to overcoming many of the related issues.
The loss is formulated as a negative log-likelihood ratio between the correct
class and all other classes, but also forces a margin between samples of
different classes.
By operating on entire class distributions instead of individual pairs or
triplets, the magnet loss potentially converges faster and leads to overall
better solutions.
%ALEX: It is good that there is literature that indicate we are onto somehting but the reviewer may ask why we didnt compare to wen2016discriminative. What makes it not suitable?
The center loss~\cite{wen2016discriminative} has been developed in an attempt
to combine classification and metric learning. % into a single objective.
The formulation utilizes a combination of a softmax classifier with an
additional term that forces compact classes by penalizing the distance of
samples to their class mean. A scalar hyperparameter balances the two
losses.
Experiments suggest that this joint formulation of classification and metric
learning produces state of the art results.

\if0
\begin{figure*}[t!]
    \centering
    \subfloat[][]{%
        \includegraphics[width=0.4\linewidth]{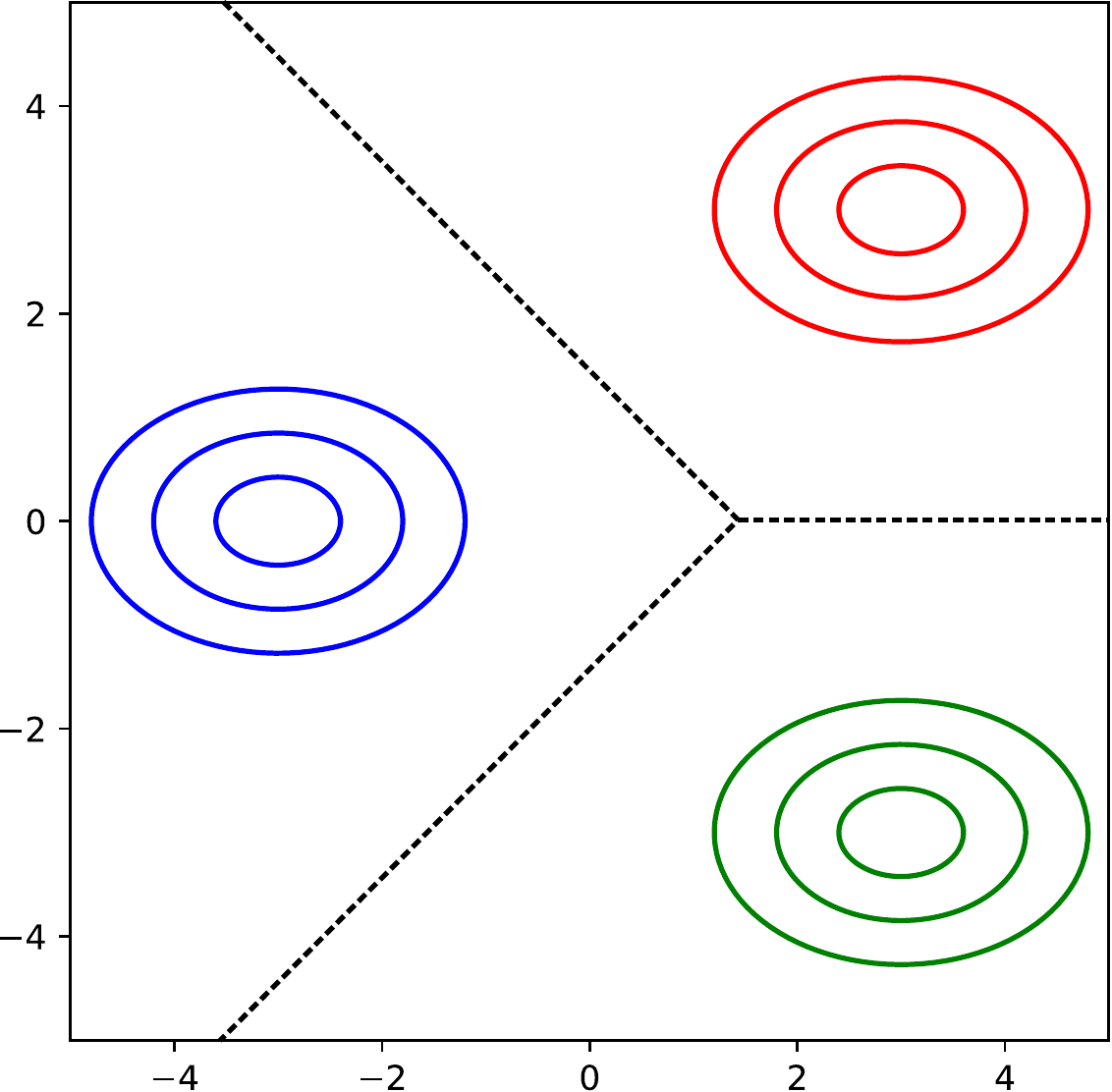}
        \label{fig:softmax-decision-a}
    }
    \hspace{2cm}
    \subfloat[][]{%
        \includegraphics[width=0.4\linewidth]{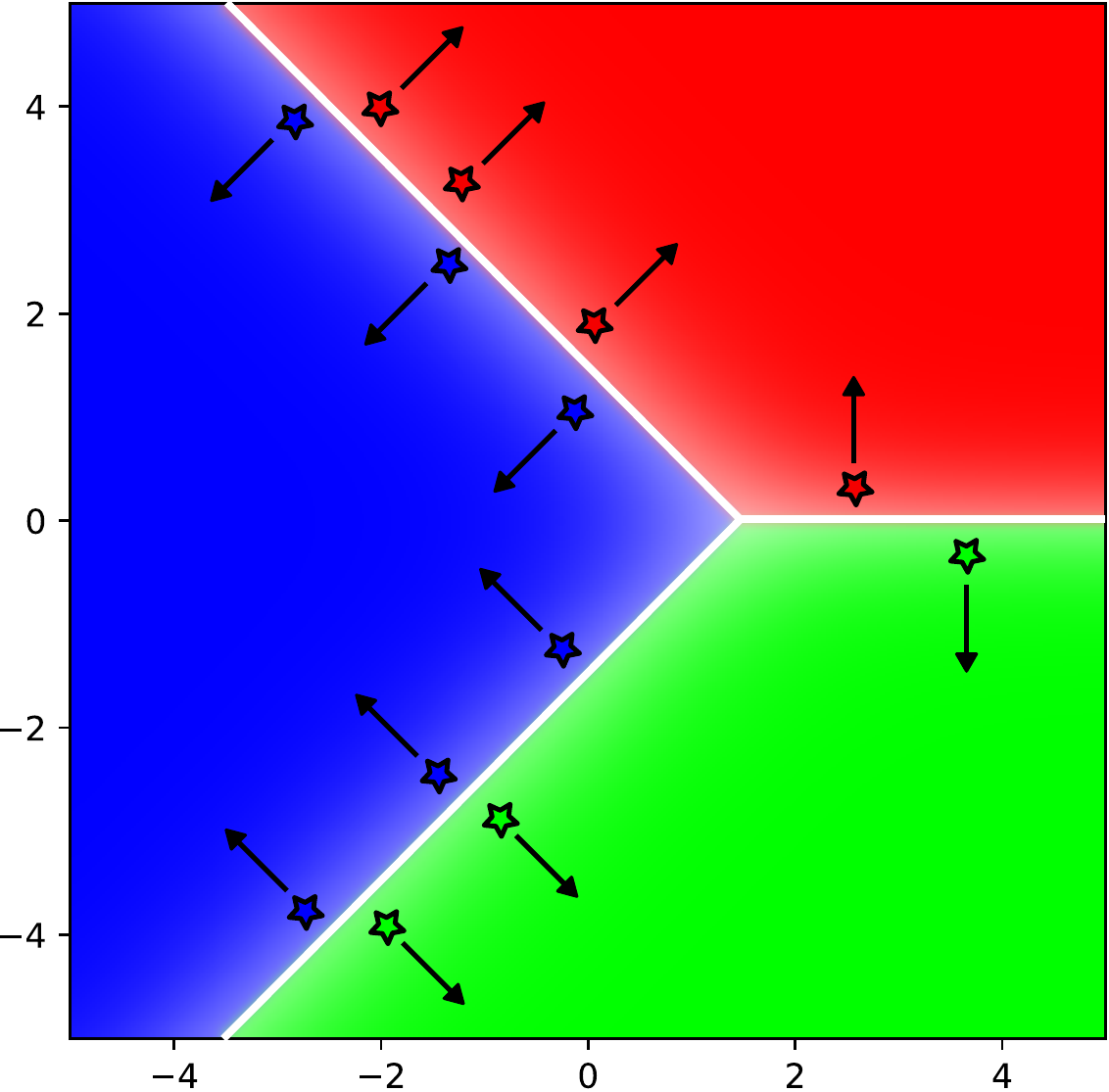}
        \label{fig:softmax-decision-b}
    }
    \caption[%
        Class-conditional densities and decision surface of a softmax
        classifier%
    ]{%
        Plot (a) shows three Gaussian class-conditional densities
        (iso-contours) and the corresponding decision boundary (dashed lines).
        Plot (b) shows the conditional class probabilities (color coded) and a
        set of hypothesized training examples.
        The softmax classifier models the posterior class probabilities
        directly, without construction of Gaussian densities.
        By training with the cross-entropy loss, samples are pushed away from
        the decision boundary, but not necessarily towards a class mean.
    }
\label{fig:softmax-decision}
\end{figure*}
\fi

\begin{figure*}[t!]
    \centering
    \subfloat[][]{%
        \includegraphics[height=5cm]{metric-standard-a}
        \label{fig:softmax-decision-a}
    }
    \hfill
    \subfloat[][]{%
        \includegraphics[height=5cm]{metric-standard-b}
        \label{fig:softmax-decision-b}
    }
    \hfill
    \subfloat[][]{%
        \includegraphics[height=5cm]{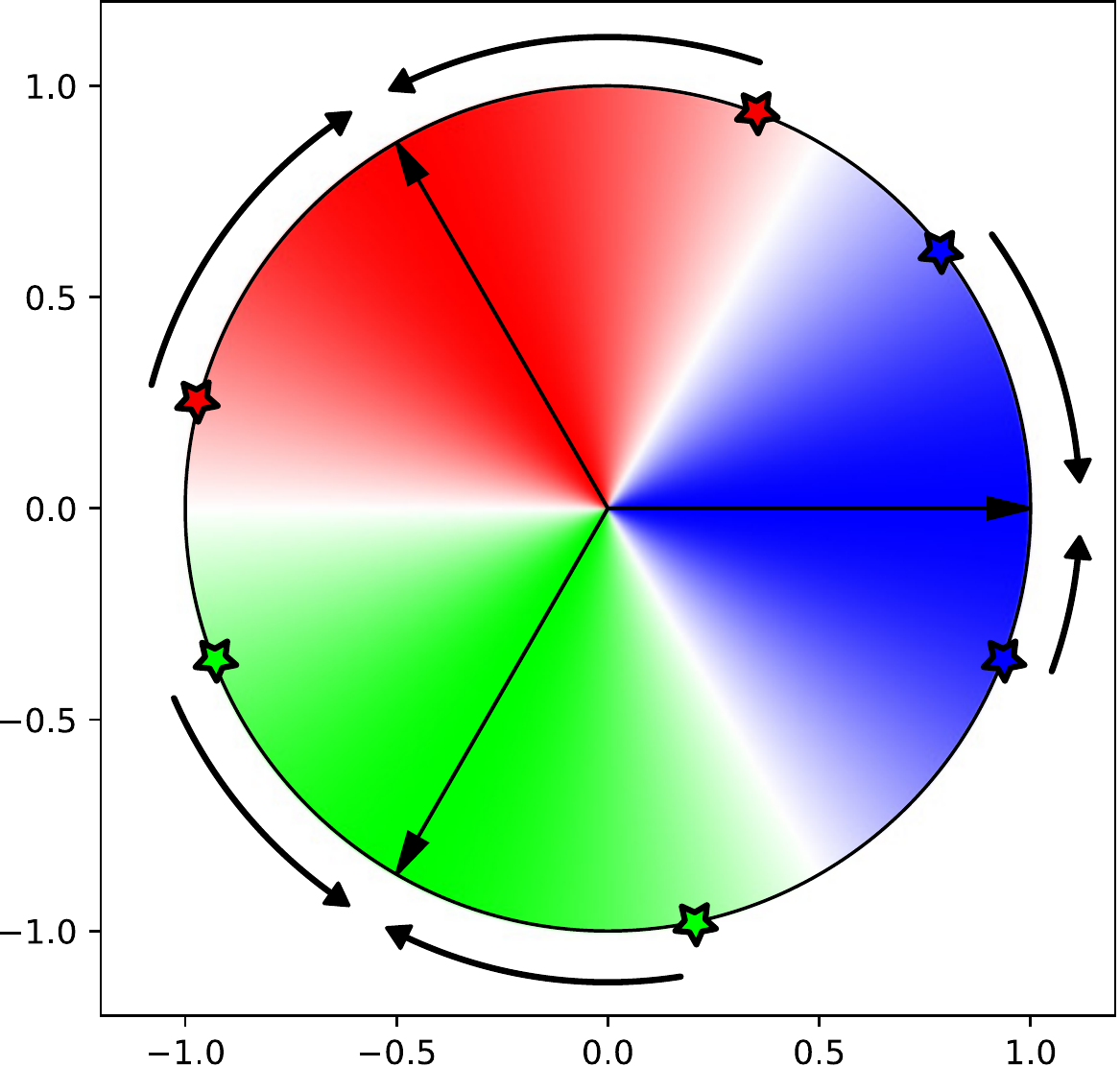}
        \label{fig:cosine-softmax}
    }
    \caption[%
        Class-conditional densities and decision surface of a softmax
        classifier%
    ]{%
        Plot (a) shows three Gaussian class-conditional densities
        (iso-contours) and the corresponding decision boundary (dashed lines).
        Plot (b) shows the conditional class probabilities (color coded) and a
        set of hypothesized training examples.
        The softmax classifier models the posterior class probabilities
        directly, without construction of Gaussian densities.
        By training with the cross-entropy loss, samples are pushed away from
        the decision boundary, but not necessarily towards a class mean.
        Plot (c) illustrates the posterior class probabilities (color coded)
        and decision boundary (white line) of the cosine softmax classifier
        for three classes.
        During training, all samples are pushed away from the decision boundary
        towards their parametrized class mean
        direction~(indicated by an arrow).
    }
\label{fig:softmax-decision}

\figurespacingSmall{}
\end{figure*}

\paragraph{Person Re-Identification}

With the availability of larger datasets, person re-identification has become
an application domain of deep metric learning and
%However,
several CNN architectures have been designed specifically for this task.
Most of them focus on mid-level features and try to deal with
pose variations and viewpoint changes explicitly by introducing special units
into the architecture.
For example, Li~\andothers~\cite{li2014deepreid} propose a CNN
with a special patch matching layer that captures the displacement between
mid-level features.
Ahmed~\andothers~\cite{ahmed2015improved} capture feature displacements
similarly by application of special convolutions that compute the difference
between neighborhoods in the feature map of two input images.
The gating functions in the network of Varior~\andothers~\cite{varior2016gated}
compare features along a horizontal stripe and output a gating mask to indicate
how much emphasis should be paid to the local patterns.
Finally, in~\cite{varior2016siamese} a recurrent siamese neural network
architecture is proposed that processes images in rows.
The idea behind the recurrent architecture is to increase contextual
information through sequential processing. % of image patches.

More recent work on person re-identification suggests that baseline
CNN architectures can compete with their specialized
counter parts.
In particular, the current best performing method on the
MARS~\cite{zheng2016mars} is a conventional residual
network~\cite{hermans2017defense}.
Application of baseline CNN architectures can be beneficial if pre-trained
models are available for fine-tuning to the person re-identification task.
Influence of pre-training on overall performance is studied
in~\cite{zheng2016mars}.
They report between $9.5\%$ and $10.2\%$ recognition rate is due to
pre-training on ImageNet~\cite{krizhevsky2012imagenet}.

\section{Standard Softmax Classifier}
\label{sec:metric-learning-standard-softmax}

\if0
\begin{figure*}[t!]
    \centering
    \subfloat[][]{%
        \includegraphics[height=6.5cm]{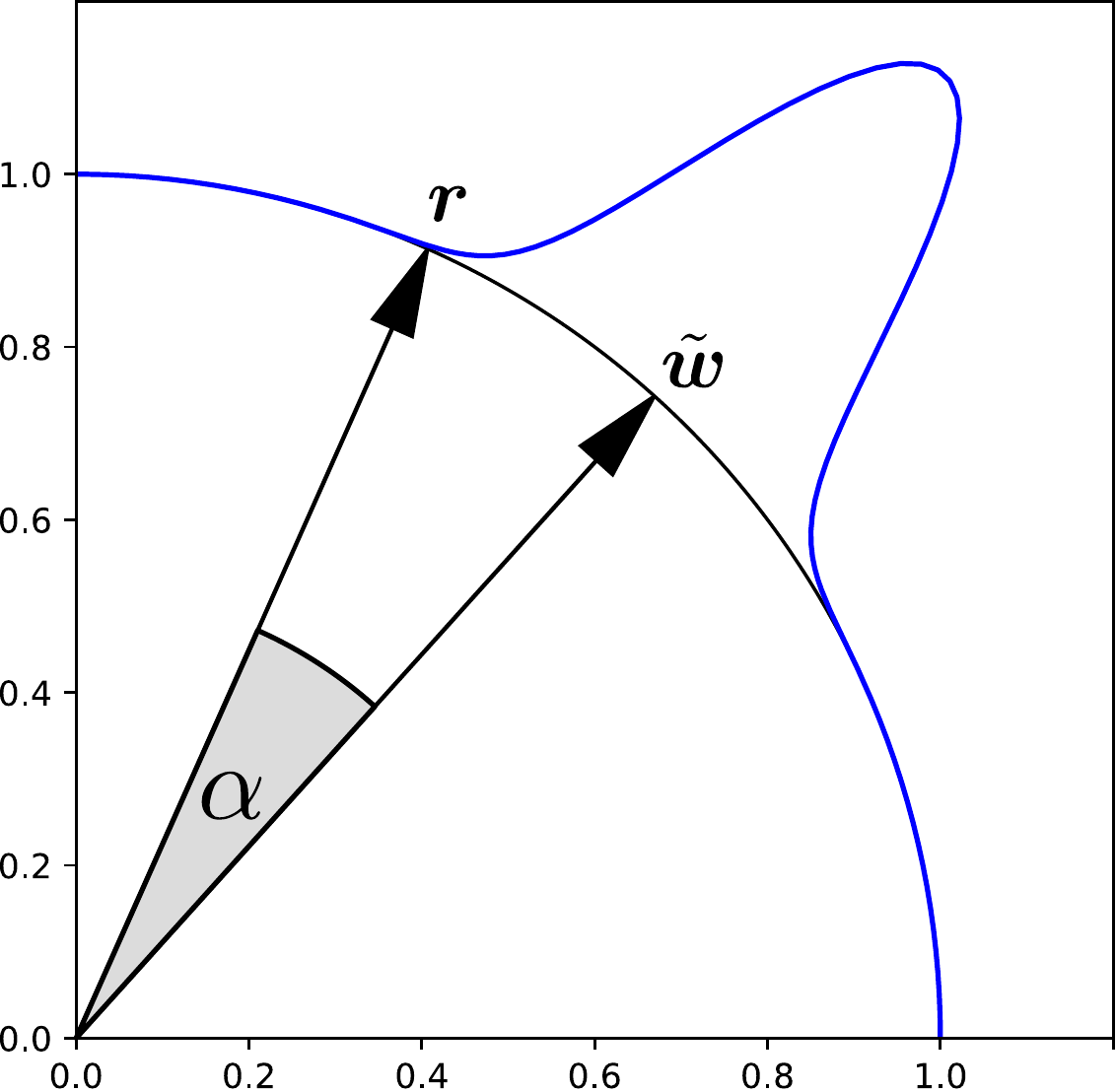}
        \label{fig:vMF}
    }
    \hspace{2cm}
    \subfloat[][]{%
        \includegraphics[height=6.5cm]{metric-cosine}
        \label{fig:cosine-softmax}
    }
    \caption[%
        Illustration of a von Mises-Fisher distribution and decision boundary
        of the cosine softmax classifier.
    ]{%
    	 \textcolor{red}{TODO: This could be combined with Figure 2 and made into a single column to save space.}
        Plot (a) illustrates a von Mises-Fisher distribution. The probability
        density increases as the cosine of the angle~$\alpha$ between
        sample~$\Feature$ and mean direction~$\tilde{\vec{w}}$ becomes
        smaller. Parameter~$\kappa$ controls the concentration of the
        distribution similar to how the standard deviation controls the
        spread of a Gaussian distribution.
        Plot (b) illustrates the posterior class probabilities (color coded)
        and decision boundary (white line) of the cosine softmax classifier
        for three classes.
        During training, all samples are pushed away from the decision boundary
        towards their parametrized class mean
        direction~(indicated by an arrow).
    }
\label{fig:vMF-and-cosine-softmax}
\end{figure*}
\fi

Given a dataset~$\Dataset = \{(\Input_i, \Label_i)\}_{i=1}^{\NumData}$
of~$\NumData$ training images~$\Input_i\in\InputSpace$ and associated class
labels~$\Label_i\in\{1, \dots, \NumLabels\}$,
the standard approach to classification in the deep learning setting is to
process input images by a CNN and place a softmax classifier on top of
the network to obtain probability scores for each of the~$\NumLabels$ classes.
The softmax classifier chooses the class with maximum probability according to
a parametric function
\begin{equation}
    p(\Label = k \mid \Feature)
    =
    \frac{%
        \exp{\left(\t{\vec{w}_k} \Feature + b_k\right)}
    }{%
        \sum_{n=1}^{\NumLabels}
        \exp{\left(\t{\vec{w}_n} \Feature + b_n\right)}
    }
    \label{eq:standard-softmax}
\end{equation}
where~$\Feature = \Encode{\Input}, \Feature\in \mathbb{R}^d$ is the underlying
feature representation
of a parametrized encoder network that is trained jointly with the classifier.
For the special case of~$\NumLabels=2$ classes this formulation is equivalent
to logistic regression.
Further, the specific choice of functional form can be motivated from a
generative perspective on the classification problem.
If the class-conditional densities are Gaussian
\begin{equation}
    \begin{split}
    p(\Feature & \| \Label = k)
    =\\&
    \frac{1}{\sqrt{|2\pi \mat{\Sigma}|}}
    \exp{\left(-\frac{1}{2}
        \t{\left(
            \Feature - \vec{\mu}_k
        \right)}
        \inv{\mat{\Sigma}}
        \left(
            \Feature - \vec{\mu}_k
        \right)
    \right)}
    \end{split}
\end{equation}
with shared covariance~$\mat{\Sigma}$, then the posterior class probability
can be computed by Bayes' rule
\begin{align}
    p(\Label = k \mid \Feature)
    &=
    \frac{%
        p(\Feature \| \Label = k)
        p(\Label = k)
    }{%
        \sum_{n=1}^{\NumLabels}
        p(\Feature \| \Label = n)
        p(\Label = n)
    }
    \\
    &=
    \frac{%
        \exp{\left(\t{\vec{w}}_k \Feature + b_k\right)}
    }{%
        \sum_{n=1}^{\NumLabels}
        \exp{\left(\t{\vec{w}}_n \Feature + b_n\right)}
    }
\end{align}
with~$
    \vec{w}_k=\inv{\mat{\Sigma}} \vec{\mu}_k
$ and~$
    b_k = -\frac{1}{2}\t{\vec{\mu}}_k\inv{\mat{\Sigma}} \vec{\mu}_k
    +
    \log p(\Label_i = k)
$~\cite{bishop2006pattern}.
However, the softmax classifier is trained in a discriminative regime.
Instead of determining the parameters of the class-conditional densities and
prior class probabilities, the parameters~$
    \{\vec{w}_1, b_1, \dots, \vec{w}_{\NumLabels}, b_{\NumLabels}\}
$ of the conditional class probabilities are obtained directly by minimization
of a classification loss.
Let~$\Indicator{\Label=k}$ denote the indicator function that evaluates to~$1$
if~$\Label$ is equal to~$k$ and~$0$ otherwise.
Then, the corresponding loss
\begin{equation}
    \LossVar(\Dataset)
    =
    -
    \sum_{i=1}^{\NumData}
    \sum_{k=1}^{\NumLabels}
    \Indicator{\Label_i = k}\cdot
    \log \, p(\Label_i = k \| \Feature_i)
    \label{eq:crossentropy-loss}
\end{equation}
% https://en.wikipedia.org/wiki/Cross_entropy
% https://www.quora.com/What-are-the-differences-between-maximum-likelihood-and-cross-entropy-as-a-loss-function
%is called cross-entropy loss because it 
minimizes the cross-entropy between
the \textit{true} label
distribution~$p(\Label = k) = \Indicator{\Label = k}$
and estimated probabilities of the softmax
classifier~$p(\Label = k \| \Feature)$.
By minimizing the cross-entropy loss, parameters are chosen such
that the estimated probability is close to~$1$ for the correct class and close
to~$0$ for all other classes.

%\figurename~\ref{fig:softmax-decision} illustrates a classification
%problem with three classes.
\figurename~\ref{fig:softmax-decision-a} shows three Gaussian
densities~$p(\Feature \| \Label)$ together with the corresponding decision
boundary.
The posterior class probabilities of this scenario are shown in
\figurename~\ref{fig:softmax-decision-b} together with a set of hypothesized
training examples.
Whereas the Gaussian densities peak around a class mean, the posterior class
probability is a function of the distance to the decision boundary.
%plane of the closest neighboring class.
When the feature encoder is trained with the classifier jointly by minimization
of the cross-entropy loss, the parameters of the encoder
network are adapted to push samples away from the decision boundary as far as
possible, but not necessarily towards the class mean that has been taken to
motivate the specific functional form. % from a generative view.
This behavior is problematic for metric learning because similarity in
terms of class membership is encoded in the orientation of the decision
boundary rather than in the feature representation itself.

\section{Cosine Softmax Classifier}
\label{sec:metric-learning-cosine-softmax}

With few adaptations the standard softmax classifier can be modified to produce
compact clusters in representation space.
First,~$\ell_2$ normalization must be applied to the final layer of the
encoder network to ensure the representation is unit length~$
    {\lVert \Encode[\EncoderParams]{\Input} \rVert}_2 = 1,
    \forall \vec{x}\in\InputSpace
$. Second, the weights must be normalized to unit-length as well, i.e.,~$
    \tilde{\vec{w}}_k
    =
    \vec{w}_k / \lVert \vec{w}_k \rVert_2,
    \forall k = 1, \dots, \NumLabels
$.
Then, the cosine softmax classifier can be stated by
\begin{equation}
    p(\Label = k \| \Feature)
    =
    \frac{%
        \exp{%%
            \left(
                \VMspread \cdot \t{\tilde{\vec{w}}_k}
                \Feature
            \right)
        }
    }{%
        \sum_{n=1}^{\NumLabels}
        \exp{%
            \left(
                \VMspread \cdot \t{\tilde{\vec{w}}_n}
                \Feature
            \right)
        }
    },
    \label{eq:modified-softmax}
\end{equation}
where~$\VMspread$ is a free scaling parameter.
This parametrization has~$\NumLabels - 1$ fewer parameters compared
to the standard formulation~(\ref{eq:standard-softmax}) because the bias
terms~$b_k$ have been removed~$
    %\ClassifierParams
    %=
    \{
        \VMspread, \tilde{\vec{w}}_1, \dots, \tilde{\vec{w}}_{\NumLabels}
    \}
$.
Otherwise, the functional form resembles
strong similarity to the standard parametrization and implementation is
straight-forward.
In particular, decoupling the length of the weight vector~$\VMspread$
from its direction has been proposed before~\cite{salimans2016weight} as a way
to accelerate convergence of stochastic gradient descent.
Training itself can be carried out using the
cross-entropy loss %~(\ref{eq:crossentropy-loss})
as usual since the cosine softmax classifier is merely a change of
parametrization compared to the standard formulation.

The functional modeling of log-probabilities
by~$\kappa \cdot \t{\tilde{\vec{w}}_k} \Feature$
can be motivated from a generative perspective as well.
If the class-conditional likelihoods follow a von Mises-Fisher (vMF)
distribution
\begin{equation}
    p(\Feature \| \Label = k)
    =
    \VMnorm
    \exp{\left(
        \VMspread\cdot
        \t{\tilde{\vec{w}}_k} \Feature
    \right)}
\end{equation}
with shared concentration parameter~$\VMspread$ and normalizer~$\VMnorm$,
then \eqname~\ref{eq:modified-softmax} is the posterior class probability
under an equal prior assumption~$
    p(\Label = k)
    =
    p(\Label = l)$,~$\forall k,l\in \{1, \dots, \NumLabels\}
$.
The vMF distribution is an isotropic probability distribution on
the~$\FeatureDim - 1$ dimensional sphere in~$\FeatureSpace$ that peaks around
mean direction~$\tilde{\vec{w}}_k$ and decays as the cosine similarity
decreases.
%This is illustrated in \figurename~\ref{fig:vMF}.

\begin{figure}[t!]
    \centering
    \subfloat[][$\VMspread=1$]{%
        \includegraphics[width=\linewidth]{%
            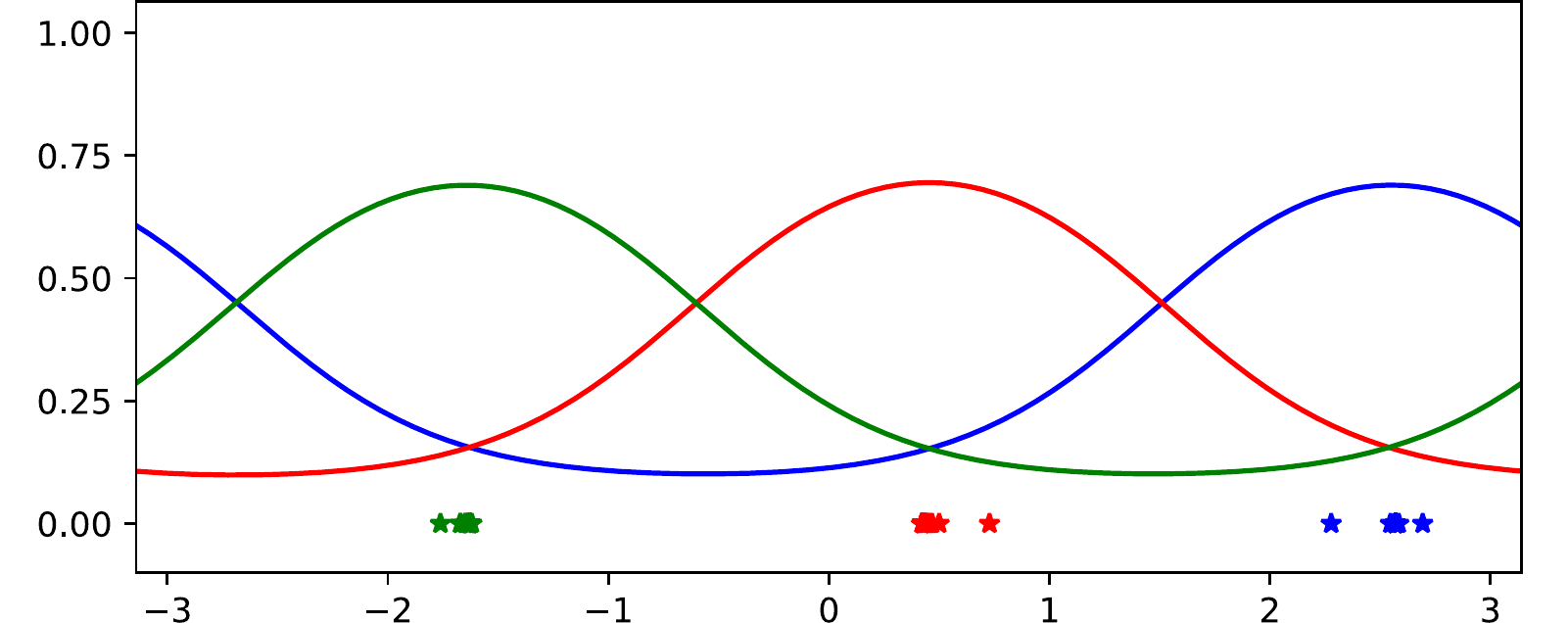}
    }
    \qquad
    \subfloat[][$\VMspread=10$]{%
        \includegraphics[width=\linewidth]{%
            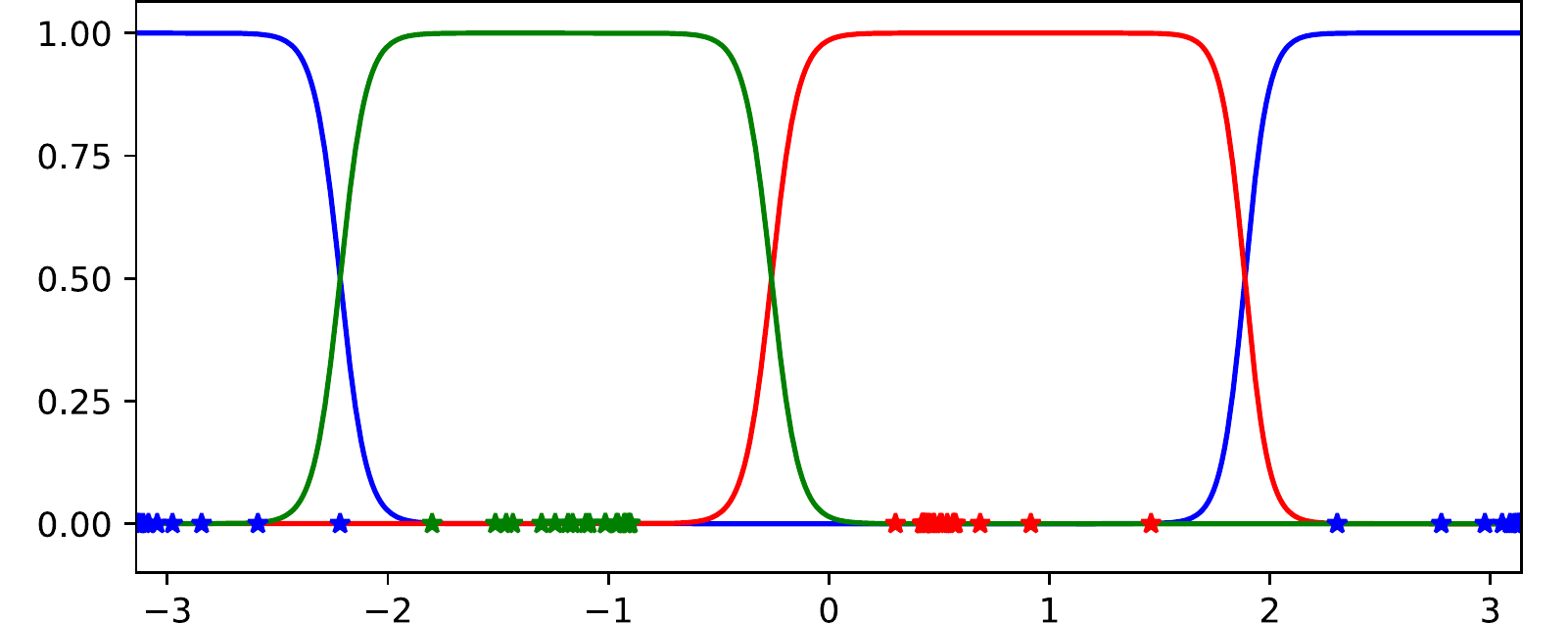}
    }
    \caption[
        Effect of free scaling parameter on conditional class probabilities in
        the cosine softmax classifier]{%
        Illustration of the free scaling parameter~$\kappa$ in a one
        dimensional problem with three classes.
        The conditional class probabilities are shown as colored functions.
        Optimized sample locations are visualized as stars at~$y=0$.
        A low~$\VMspread$ value (a) leads to smoother functions with wider
        support, such that samples are pushed into tight clusters.
        The shape becomes box-like for high values~(b), allowing samples to
        move more freely within a region that is occupied by the class.
    }
    \label{fig:cosine-softmax-centration}

\figurespacing{}
\end{figure}

To understand why this parametrization enforces a cosine similarity on the
representation space, observe that the log-probabilities are directly
proportional to the cosine similarity between training examples and a
parametrized class mean direction.
By minimizing the cross-entropy loss, examples are pushed away from the
decision boundary towards their parametrized mean as
illustrated in \figurename~\ref{fig:cosine-softmax}.
In consequence, parameter vector~$\tilde{\vec{w}}_k$ becomes a surrogate for
all samples in cases~$k$.
The scaling parameter~$\VMspread$ controls the shape of the conditional class
probabilities as
illustrated in \figurename~\ref{fig:cosine-softmax-centration}.
A low value corresponds to smoother functions with wider support.
A high~$\VMspread$ value leads to conditional class probabilities that are
box-like shaped around the decision boundary.
This places a larger penalty on misclassified examples, but at the same time
leaves more room for samples to move freely in the region of representation
space that is occupied by its corresponding class.
In this regard, the scale takes on a similar role to margin parameters in
direct metric learning objectives.
When the scale is left as a free parameter, the optimizer gradually increases
its value as the overlap between classes reduces.
A margin between samples of different classes can be enforced by regularizing
the scale with weight decay.

\section{Evaluation}
\label{sec:metric-learning-evaluation}

The first part of the evaluation compares both the training behavior and
validation error between our loss formulation and common metric learning losses
using a network trained from scratch. In the second part, overall system
performance is established against existing re-identification systems on the
same datasets.

% Alternative:
%The first part of the following evaluation focuses on the effectiveness of the
%proposed loss formulation.
%We repeatedly train a network from scatch using different loss formulations and
%compare training behavior and validation error.
%In the second part, overall system performance is established against existing
%re-identification systems on the same datasets.

\subsection{Network Architecture}
\label{sec:metric-learning-architecture}

%This section presents a light-weight CNN architecture for person
%re-identification and people tracking.
The network architecture used in our experiments is relatively shallow to allow
for fast training and inference, e.g., for application in the related task
of appearance based object tracking~\cite{blind0000ref}.
The architecture is summarized in \tablename~\ref{tab:network-architecture}.
%The network used in our experiments matches the architecture used in the
%related task of appearance based object tracking~\cite{blind0000ref} and is summarized in
%\tablename~\ref{tab:network-architecture}.
Input images are rescaled to~$128\times 64$ and presented to the network in
RGB color space.
A series of convolutional layers reduces the size of the feature map
to~$16\times 8$ before a global feature vector of length~$128$ is extracted
by layer \textit{Dense 10}.
The final~$\ell_2$ normalization projects features onto the unit
hypersphere for application of the cosine softmax classifier.
The network contains several residual blocks that follow
the pre-activation layout proposed by He~\andothers~\cite{he2016identity}.
%In general, a residual block maps the input~$\vec{x}$ to an
%output~$\vec{y}$ by~$\vec{y} = h(i(\vec{x}) + g(\vec{x}))$ where~$i(\vec{x})$
%is an identity mapping,~$g(\vec{x})$ is a residual function and~$h(\vec{x})$
%is a nonlinearity.
%In the pre-activation residual block, all non-linearities are moved into
%the residual function such that there is always a direct flow of information
%between the input and output, i.e.,~$\vec{y} = i(\vec{x}) + g(\vec{x})$.
%This has shown to ease learning and improve
%overall accuracy~\cite{he2016identity}.
%The structure of the residual blocks is shown in
%\figurename~\ref{fig:preactivation-residual}.
The design follows the ideas of wide residual
networks~\cite{zagoruyko2016wide}:
All convolutions are of size~$3\times 3$ and max pooling is replaced by
convolutions of stride~$2$.
When the spatial resolution of the feature map is reduced, then the number of
channels is increased accordingly to avoid a bottleneck.
Dropout and batch normalization are used as means of regularization.
Exponential linear units~\cite{clevert2015fast} are used as activation
function in all layers.

Note that with in total 15 layers (including two convolutional layers in each
residual block) the network is relatively shallow when compared to the current
trend of ever deeper architectures~\cite{he2016identity}.
This decision was made for the following two reasons.
First, the network architecture has been designed for the application of both
person re-identification and online people tracking~\cite{blind0000ref},
where the latter requires fast computation of appearance features.
In total, the network has 2,800,864 parameters and one forward pass
of 32 bounding boxes takes approximately~$30\,\textrm{ms}$ on an Nvidia GeForce
GTX 1050 mobile GPU{}. Thus, this network is well suited for online tracking
even on low-cost hardware.
%This decision has been made with the people tracking application in mind, to
%permit application in online tracking scenarios, and is backed up by the
%experiments of~\cite{zagoruyko2016wide} that suggest shallow networks with
%increased width can be computationally more efficient and achieve
%similar accuracy.
Second, architectures that have been designed for person
re-identification specifically~\cite{li2014deepreid,ahmed2015improved}
put special emphasis on mid-level features.
Therefore, the dense layer is added at a point where the feature map still
provides \textit{enough} spatial resolution.

\begin{table}[t!]
    \begin{tabular}{p{0.27\linewidth}cc}
        \toprule
        \textbf{Name} & \textbf{Patch Size/Stride} & \textbf{Output Size} \\
        \midrule
        Conv 1 & $3\times 3$/$1$ & $32\times 128\times 64$ \\
        Conv 2 & $3\times 3$/$1$ & $32\times 128\times 64$ \\
        Max Pool 3 & $3\times 3$/$2$ & $32\times 64\times 32$ \\
        Residual 4 & $3\times 3$/$1$ & $32\times 64\times 32$ \\
        Residual 5 & $3\times 3$/$1$ & $32\times 64\times 32$ \\
        Residual 6 & $3\times 3$/$2$ & $64\times 32\times 16$ \\
        Residual 7 & $3\times 3$/$1$ & $64\times 32\times 16$ \\
        Residual 8 & $3\times 3$/$2$ & $128\times 16\times 8$ \\
        Residual 9 & $3\times 3$/$1$ & $128\times 16\times 8$ \\
        Dense 10 & & $128$ \\
        \multicolumn{2}{l}{$\ell_2$ normalization} & $128$ \\
        \bottomrule
    \end{tabular}
    \caption[Overview of the person re-identification CNN architecture]{%
        Overview of the CNN architecture. The final $\ell_2$
        normalization projects features onto the unit hypersphere.}
\label{tab:network-architecture}

\figurespacing{}
\end{table}

\subsection{Datasets and Evaluation Protocols}

Evaluation is carried out on the Market~1501~\cite{zheng2015scalable}
and MARS~\cite{zheng2016mars}.
Market~1501 contains 1,501 identities and roughly 30,000 images taken from six
cameras. MARS is an extension of Market 1501 that contains 1,261 identities and
over 1,100,000 images.
The data has been generated using a multi-target tracker that generates
tracklets, i.e.\ short-term track fragments, which have then been manually
annotated to consistent identities.
Both datasets contain considerate bounding box misalignment and labeling
inaccuracies.
%The two datasets have been selected for evaluation are large enough to train
%a CNN from scratch.
For all experiments a single-shot, cross-view evaluation protocol is adopted,
i.e. a single query image from one camera is matched against a gallery of
images taken from different cameras.
The gallery image ranking is established using cosine similarity
or Euclidean distance, if appropriate.
Training and test data splits are provided by the authors.
Additionally,~10\% of the training data is split for hyperparameter tuning
and early stopping.
On both datasets cumulative matching characteristics~(CMC) at rank~$1$ and~$5$
as well as mean average precision~(mAP) are reported.
%The CMC rank~$k$ metric reports the frequency by which a matching image is
%contained in the top~$k$ images of the ranked gallery.
%If the precision at~$k$ is the number of correct items in the ranked gallery
%among all top~$k$ items, then average precision is the average precision over
%all~$k$ and mean average precision is the mean of average precision over all
%queries.
%For both metrics higher values indicate better peformance.
The scores are computed with evaluation software provided by the corresponding
dataset authors.

\if0
\paragraph{Market 1501}
The Market~1501~\cite{zheng2015scalable} dataset contains 1,501 identities and
roughly 30,000 images taken from six cameras.
This dataset is challenging due to frequent bounding box misalignment and
additional false alarms in the test gallery.
The standard evaluation protocol proposed for this dataset is followed here:
Training and testing are carried out on provided data splits (751 identities
for training, 750 for testing) using images that have been generated by a
deformable parts model detector.
Additional distractor training images that are also provided by the authors
are not used.

\paragraph{MARS}
The MARS~\cite{zheng2016mars} dataset is an extension of Market 1501 that
contains 1,261 identities and over 1,100,000 images.
The data has been generated using a multi-target tracker that generates
tracklets, i.e.\ short-term track fragments, which have then been manually
annotated to consistent identities.
Consequently, this dataset also contains significant bounding box misalignment
and inaccurate labelling.
\fi

\subsection{Baseline Methods}

In order to assess the performance of the joint classification and metric
learning framework on overall performance, the network architecture is
repeatedly trained with two baseline direct metric learning objectives.

\paragraph{Triplet loss}

The triplet loss~\cite{weinberger2009distance} is defined over tuples of
three examples~$\Feature_a$,~$\Feature_p$, and~$\Feature_n$ that include a
positive pair~$\Label_a = \Label_p$ and a negative
pair~$\Label_a \neq \Label_n$.
For each such triplet the loss demands that the difference of the distance
between the negative and positive pair is larger than a pre-defined
margin~$m\in \mathbb{R}$:
\begin{equation}
    \LossVar_{\text{t}}
    (\Feature_a, \Feature_p, \Feature_n)
    =
    {\left\{
        {\lVert \Feature_a - \Feature_n\rVert}_2
        \! - \!
        {\lVert \Feature_a - \Feature_p\rVert}_2
        \! + \!
        m
    \right\}}_{+},
\end{equation}
where~${\left\{\right\}}_{+}$ denotes the hinge function that evaluates
to~$0$ for negative values and identity otherwise.
In this experiment, a soft-margin version of the original triplet
loss~\cite{hermans2017defense} is used where the hinge is replaced by a
soft plus function~${\left\{x + m\right\}}_{+} = \log(1 + \exp(x))$ to avoid
issues with non-smoothness~\cite{rippel2015metric}.
Further, the triplets are generated directly on GPU as proposed
by~\cite{hermans2017defense} to avoid potential issues in the sampling
strategy.
Note that this particular triplet loss formulation has been used to train the
current best performing model on the MARS{} %~\cite{zheng2016mars}
dataset.

\paragraph{Magnet loss}

The magnet loss has been proposed as an alternative to siamese loss
formulations that works on entire class distribution rather than individual
samples. The loss is a likelihood ratio measure that forces separation in terms
of each sample's distance away from the means of other classes.
In its original proposition~\cite{rippel2015metric} the loss
takes on a multi-modal form.
Here, a simpler, unimodal variation of this loss is employed as it better
fits the single-shot person re-identification task:
\begin{equation}
    \LossVar_{\text{m}}
    (\Label, \Feature)
    =
    {\left\{
    -
    \log
    \frac{%
        %\exp{\left( -\frac{1}{2\hat{\sigma}^2}
        \operatorname{e}^{-\frac{1}{2\hat{\sigma}^2}
        {\lVert
            \Feature - \hat{\vec{\mu}}_{\Label}
        \rVert}_2^2
        - m
        }
        %\right)}
    }{%
        \sum_{k\in \bar{\mathcal{\NumLabels}}(\Label)}
        %\smashoperator{\sum\limits_{\substack{k\in \{1, \dots, \NumLabels\}\\k\neq \Label}}}
        %\exp{\left( -\frac{1}{2\hat{\sigma}^2}
        \operatorname{e}^{-\frac{1}{2\hat{\sigma}^2}
        {\lVert
            \Feature - \hat{\vec{\mu}}_{k}
        \rVert}_2^2
        }
        %\right)}
    }
    \right\}}_{+},
\end{equation}
where~$
    \bar{\mathcal{\NumLabels}}(y)
    =
    %\{1, \dots, \Label - 1, \Label + 1, \dots, \NumLabels\}
    \{1, \dots, \NumLabels\} \setminus \{y\}
$,~$m$ is again a margin parameter,~$\hat{\vec{\mu}}_{\Label}$ is the sample
mean of class~$\Label$, and~$\hat{\sigma}^2$ is the variance of all
samples away from their class mean. These parameters are computed on GPU
for each batch individually.

\subsection{Results}

\begin{figure}[t!]
    \subfloat[][Evolution of validation set accuracy]{%
        \includegraphics[width=\linewidth]{%
            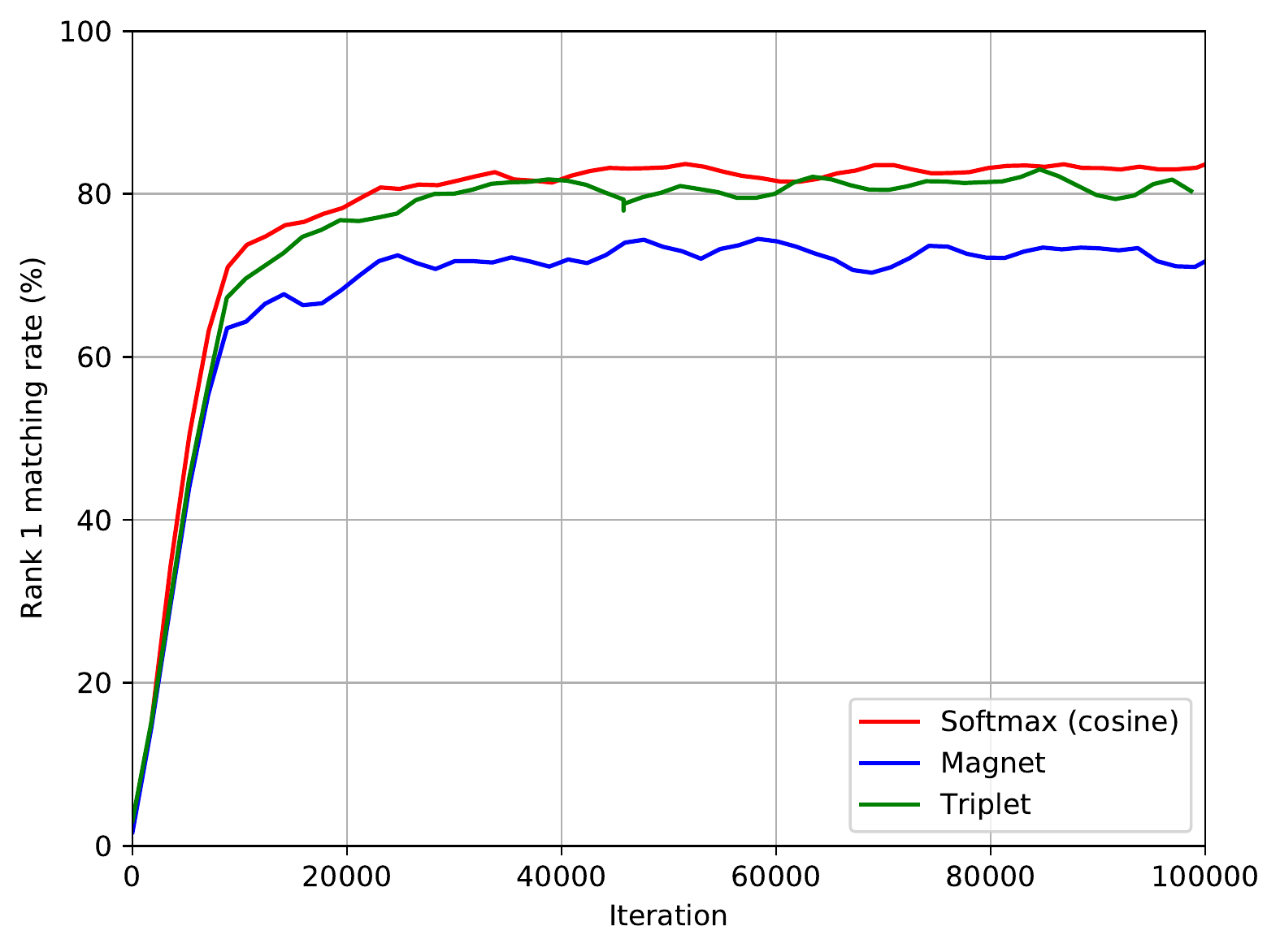}
        \label{fig:metric-learning-validation-loss}
    }
    \quad
    \subfloat[][Evolution of triplet loss on training set]{%
        \includegraphics[width=\linewidth]{%
            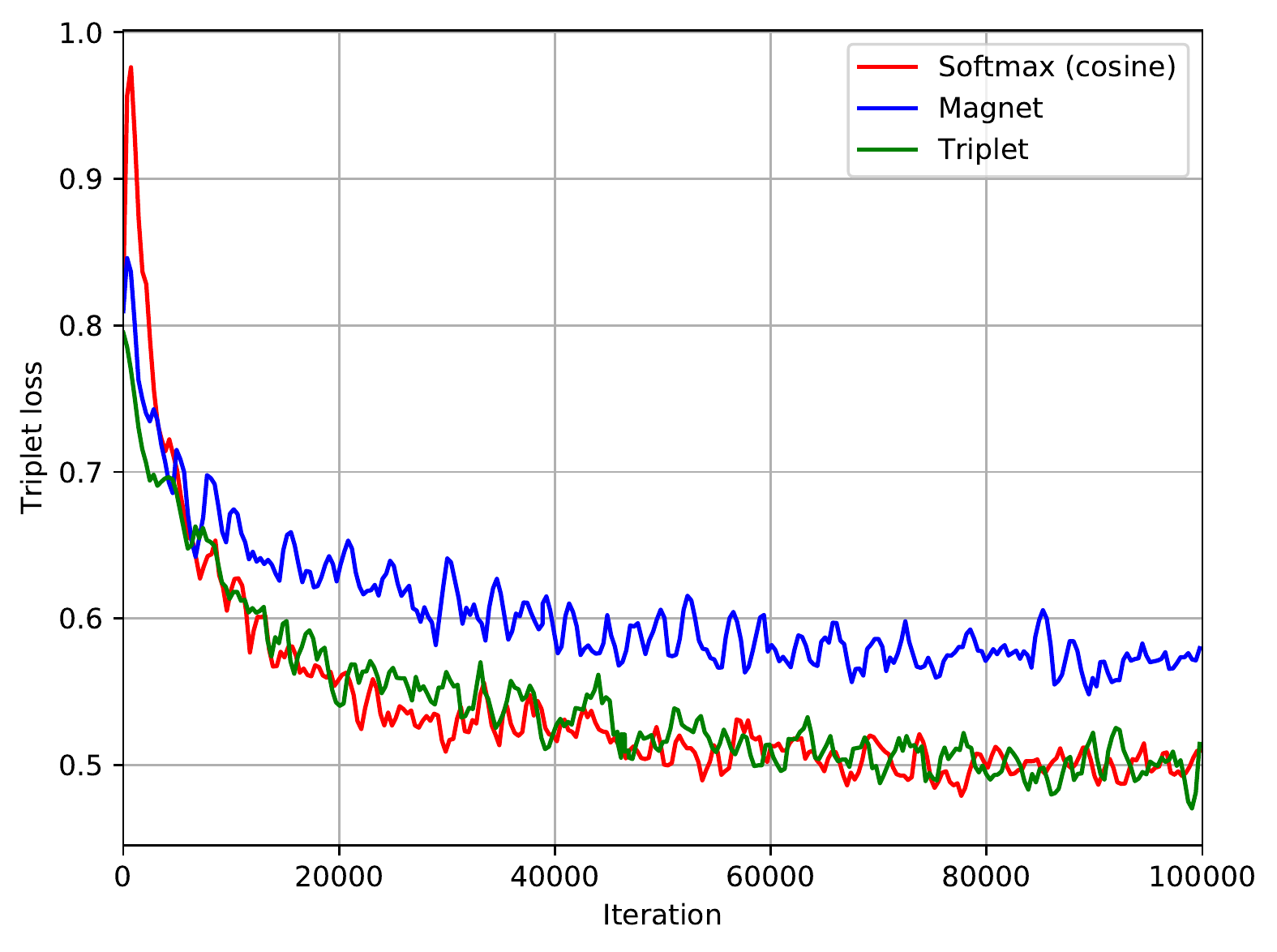}
        \label{fig:metric-learning-triplet-loss}
    }
    \caption[Evolution of validation accuracy and triplet loss]{%
        Plot (a) shows the rank 1 matching accuracy on the validation set
        as a function of training iterations.
        Plot~(b) shows how the triplet loss evolves on the training set.
        Note that the triplet loss is only used as training objective for
        the triplet network.
        For the other two methods the loss is only monitored to obtain insight
        into the training behavior.
    }
\label{fig:metric-learning-loss-plots}

\figurespacing{}
\end{figure}

The results reported in this section have been established by training the
network for a fixed number of~$100,000$ iterations using
Adam~\cite{kingma2014adam}.
The learning rate was set to~$\num{1e-3}$.
As can be seen in \figurename~\ref{fig:metric-learning-loss-plots} all
configurations have fully converged at this point.
The network was regularized with a weight decay of~$\num{1e-8}$ and dropout
inside the residual units with probability~$0.4$.
The margin of the magnet loss has been set to~$m=1$ and the cosine softmax
scale~$\VMspread$ was left as a free parameter for the optimizer to
tune, but regularized with a weight decay of~$\num{1e-1}$.
The batch size was fixed to~$128$ images.
Gallery rankings are established using Euclidean distance in case of
magnet and triplet loss, while cosine similarity is used for the softmax
classifier.
To increase variability in the training set, input images have been randomly
flipped, but no random resizing or cropping has been performed.

\paragraph{Training Behavior}

\figurename~\ref{fig:metric-learning-validation-loss} shows the rank 1
matching rate on the validation set of MARS %~\cite{zheng2016mars}
as a function of training iterations.
The results obtained on Market 1501 %~\cite{zheng2015scalable}
are omitted
here since the training behavior is similar.
The network trained with cosine softmax classifier achieves overall
best performance, followed by the network trained with soft-margin triplet
loss.
The best validation performance of the softmax network is reached at
%iteration~$86815$ with rank 1 matching rate~$0.8419$.  % this is smoothed and used for test.
iteration~$49\,760$ with rank~1 matching rate~$84.92\%$.  % top validation performer.
The best performance of the triplet loss network is reached at
iteration~$86\,329$ with rank~1 matching rate~$83.23\%$.
The magnet loss network reaches its best performance at iteration~$47\,677$
with rank~1 matching rate~$77.34\%$.
Overall, the convergence behavior of the three losses is similar, but the
magnet loss falls behind on final model performance.
In its original implementation~\cite{rippel2015metric} the authors sample
batches such that similar classes appear in the same batch.
For practical reasons such more informative sample mining has not been
implemented.
Instead, a fixed number of images per individual was randomly selected for
each batch.
Potentially, the magnet loss suffers from this less informative sampling
strategy more than the other two losses.

During all runs the triplet loss has been monitored as an additional
information source on training behavior.
\figurename~\ref{fig:metric-learning-triplet-loss} plots the triplet loss
as a function of training iterations.
Note that the triplet loss has not been used as a training objective in runs
softmax (cosine) and magnet. Nevertheless, both minimize the triplet loss
indirectly.
In particular the softmax classifier is quite efficient at minimizing the
triplet loss.
During iterations~20,000 to~40,000 the triplet loss drops even slightly faster
when optimization is carried out with the softmax classifier rather than
optimizing the triplet loss directly.
Therefore, the cosine softmax classifier effectively enforces a similarity
metric onto the representation space.

\paragraph{Re-Identification Performance}

\begin{table}[t!]
    \newcommand{\RankN}[1]{Rank #1}
    \newcommand{\C}[1]{\multicolumn{1}{c}{#1}}
    \definecolor{SeqColor}{rgb}{0.9,0.9,0.9}
    \centering
    \begin{tabular}{%
            >{\quad}
            p{0.45\linewidth}
            >{\RaggedLeft\arraybackslash}p{0.05\linewidth}
            >{\RaggedLeft\arraybackslash}p{0.05\linewidth}
            >{\RaggedLeft\arraybackslash}p{0.05\linewidth}
            }
        \toprule
        \multirow{2}{*}{\textbf{Method}} &
            \multicolumn{3}{c}{\textbf{Market 1501}} \\
        &
            \C{\textbf{\RankN{1}}} & \C{\textbf{\RankN{5}}} & \textbf{mAP} \\
        \midrule
        TriNet~\cite{hermans2017defense}\textsuperscript{a,b} & \C{\SB{84.92}} & \C{94.21} & \C{\SB{69.14}} \\
        LuNet~\cite{hermans2017defense}\textsuperscript{b} & \C{81.38} & \C{92.34} & \C{60.71} \\
        IDE + XQDA~\cite{zheng2016mars}\textsuperscript{a,\textdagger} & \C{73.60} & \C{-} & \C{49.05} \\
        DaF~\cite{yu2017divide}\textsuperscript{a} & \C{82.30} & \C{-} & \C{\BS{72.42}} \\
        JLML~\cite{li2017person}\textsuperscript{a} & \C{\BS{85.10}} & \C{-} & \C{65.50} \\
        GoogLeNet~\cite{zhao2017deeply}\textsuperscript{a} & \C{81.00} & \C{-} & \C{63.40} \\
        SVDNet~\cite{sun2017svdnet}\textsuperscript{a} & \C{82.30} & \C{-} & \C{62.10} \\
        Gated CNN~\cite{varior2016gated}\textsuperscript{b} & \C{65.88} & \C{-} & \C{39.55} \\
        Recurrent CNN~\cite{varior2016siamese}\textsuperscript{b} &  \C{61.60} & \C{-} & \C{35.30} \\
        \midrule
        Ours (triplet)\textsuperscript{b} & \C{\SB{74.88}} & \C{\SB{88.72}} & \C{\SB{53.04}} \\
        Ours (magnet) & \C{61.10} & \C{81.03} & \C{40.12} \\
        Ours (cosine softmax) & \C{\BS{79.10}} & \C{\BS{91.06}} & \C{\BS{56.68}} \\
        \bottomrule
    \end{tabular}
    \caption[%
        Person re-identification results on Market 1501]{%
        Performance comparison on Market 1501~\cite{zheng2015scalable}.
        \textsuperscript{\textdagger}: Numbers taken
        from~\cite{hermans2017defense}.
        Methods below the line show our network architecture trained with
        different losses.
        \textsuperscript{a}: Pre-trained on ImageNet.
        \textsuperscript{b}: Siamese network.}
    \label{tab:metric-learning-results-market}

\figurespacing{}
\end{table}

All three networks have been evaluated on the provided test splits of the
Market 1501 %~\cite{zheng2015scalable}
and MARS %~\cite{zheng2016mars}
datasets.
\tablename~\ref{tab:metric-learning-results-market}
and~\ref{tab:metric-learning-results-mars} summarize the results and
provide a comparison against the state of the art.
The training behavior and rank 1 matching rates that have been observed on the
validation set manifest in the final performance on the provided test splits.
Of our own networks, on both datasets the cosine softmax network achieves the
best results, followed by the siamese network.
The gain in mAP due to the softmax loss is~$3.64$ on the Market 1501 dataset
and~$2.58$ on the MARS dataset.
This is a relative gain of~$6.8\%$ and~$4.7\%$ respectively.
The state of the art contains several alternative siamese architectures that
have been trained with a contrastive or triplet loss, marked
by~\textsuperscript{b} in \tablename~\ref{tab:metric-learning-results-market}
and~\ref{tab:metric-learning-results-mars}.
The performance of these networks is not always directly comparable, because
the models have varying capacity.
However, the LuNet of Hermans~\andothers~\cite{hermans2017defense} is a
residual network with roughly double the capacity of the proposed architecture.
The reported numbers have been generated with test-time data augmentation that
accounts for approximately 3 mAP points according to the corresponding authors.
Thus, the proposed network %trained with cosine softmax classifier
comes in close range at much lower capacity.
Further, the method of~\cite{zheng2016mars} refers to a CaffeNet that has
been trained with the conventional softmax classifier and the metric subspace
has been obtained in a separate post processing step.
The results suggest that the proposed joint classification and metric learning
framework %of \sectionname~\ref{sec:metric-learning-joint}
not only enforces a metric onto the representation space, but also that
encoding the metric directly into the classifier works better than treating it
in a subsequent post processing step.

%Due to the targeted tracking application, the proposed network architecture is
%much smaller than most of the architectures employed in the state of the art.
The best performing method on Market 1501 has a~$15.84$ points higher mAP score
than the cosine softmax network. On MARS, the
best performing method achieves a~$10.82$ higher mAP.
This is a large-margin improvement over the proposed network, which shows that
considerate improvement is possible by application of larger capacity
architectures with additional pre-training.
Note that, for example, the TriNet~\cite{hermans2017defense} is a
ResNet-50~\cite{he2016deep} with~$25.74$ million parameters that has been
pre-trained on ImageNet~\cite{krizhevsky2012imagenet}.
With roughly a tenth of the parameters, our network has much lower capacity.
%Clearly, deeper networks have an advantage over the small architecture used
%in this work, but they are not applicable in online tracking scenarios due to
%the computational resources that they require.
The best performing network that has been trained from scratch, i.e.,
without pre-training on ImageNet, is the LuNet of
Hermans~\andothers~\cite{hermans2017defense}.
With approximately 5 million parameters the network is still roughly double
the size, % of the proposed architecture,
but the final model performance in
terms of mAP is only~$4.03$ and~$3.6$ points higher (including test-time
augmentation). % than the proposed architecture.
Therefore, %for the targeted tracking application
the proposed architecture provides a good trade off between computational
efficiency and re-identification performance.

\begin{table}[t!]
    \newcommand{\RankN}[1]{Rank #1}
    \newcommand{\C}[1]{\multicolumn{1}{c}{#1}}
    \definecolor{SeqColor}{rgb}{0.9,0.9,0.9}
    \centering
    \begin{tabular}{%
            >{\quad}
            p{0.45\linewidth}
            >{\RaggedLeft\arraybackslash}p{0.05\linewidth}
            >{\RaggedLeft\arraybackslash}p{0.05\linewidth}
            >{\RaggedLeft\arraybackslash}p{0.05\linewidth}
            >{\RaggedLeft\arraybackslash}p{0.05\linewidth}
            >{\RaggedLeft\arraybackslash}p{0.05\linewidth}
            >{\RaggedLeft\arraybackslash}p{0.05\linewidth}
            }
        \toprule
        \multirow{2}{*}{\textbf{Method}} &
            \multicolumn{3}{c}{\textbf{MARS}} \\
        &
            \C{\textbf{\RankN{1}}} & \C{\textbf{\RankN{5}}} & \textbf{mAP} \\
        \midrule
        TriNet~\cite{hermans2017defense}\textsuperscript{a,b} & \C{\BS{79.80}} & \C{91.36} & \C{\BS{67.70}} \\  % pre-trained
        LuNet~\cite{hermans2017defense}\textsuperscript{b} & \C{\SB{75.56}} & \C{89.70} & \C{\SB{60.48}} \\   % from scatch (-3% mAP for test-time augmentation)
        IDE + XQDA~\cite{zheng2016mars}\textsuperscript{a,\textdagger} & \C{65.30} & \C{82.00} & \C{47.60} \\
        MSCAN~\cite{li2017learning} & \C{71.77} & \C{86.57} & \C{56.06} \\ %
        P-QAN~\cite{liu2017quality} & \C{73.73} & \C{84.90} & \C{51.70} \\ % set to set, 
        CaffeNet~\cite{zhou2017see} & \C{70.60} & \C{90.00} & \C{50.70} \\ % similar to AlexNet, with spatial consistency
        \midrule
        Ours (triplet)\textsuperscript{b} & \C{\SB{71.31}} & \C{\SB{85.55}} & \C{\SB{54.30}} \\
        Ours (magnet) & \C{63.13} & \C{81.16} & \C{45.45} \\
        Ours (cosine softmax) & \C{\BS{72.93}} & \C{\BS{86.46}} & \C{\BS{56.88}} \\
        \bottomrule
    \end{tabular}
    \caption[%
        Person re-identification results on MARS]{%
        Performance comparison on MARS~\cite{zheng2016mars}.
        Methods below the line show our network architecture trained with
        different losses.
        \textsuperscript{\textdagger}: Numbers taken
        from~\cite{hermans2017defense}.
        \textsuperscript{a}: Pre-trained on ImageNet.
        \textsuperscript{b}: Siamese network.}
    \label{tab:metric-learning-results-mars}

\figurespacing{}
\end{table}

\paragraph{Learned Embedding}

\begin{figure}[h]
    %\centering
    %\subfloat[][Examples for successful queries without error in top~5 matches]{%
    %    \includegraphics[width=\linewidth]{good_2}
    %    \label{fig:market-query-good}
    %}
    %\qquad
    %\subfloat[][Interesting failure cases]{%
        \includegraphics[width=\linewidth]{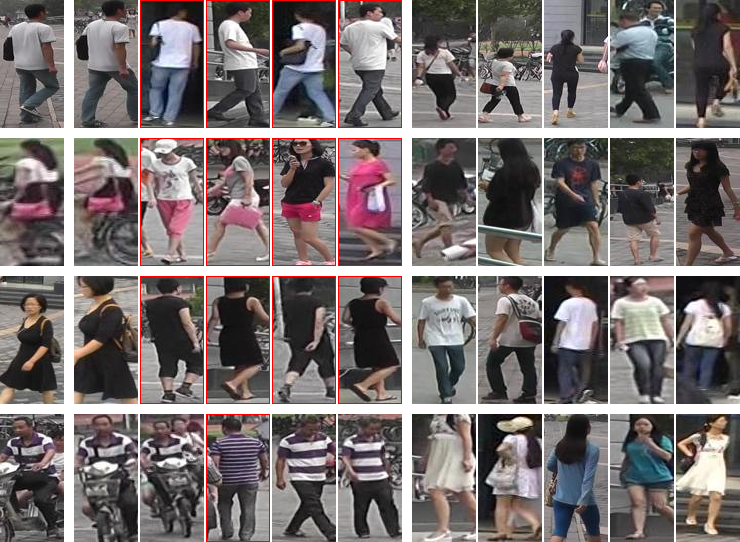}
    %    \label{fig:market-query-bad}
    %}
    \caption[Example queries from Market~1501 test set]{%
        Failure cases on example queries generated from
        Market~1501~\cite{zheng2015scalable} test gallery.
        For each query image the five most similar and dissimilar images
        are shown.
        %The first image in
        %each row shows the query image. The second block shows the five most
        %similar images in the gallery (errors marked by red border).
        %The third block shows the five most dissimilar images in the gallery.
    }
    %\label{fig:market-queries}
    \label{fig:market-query-bad}

%\figurespacing{}
\end{figure}

\begin{figure}[h]
    \centering
    \includegraphics[width=\linewidth]{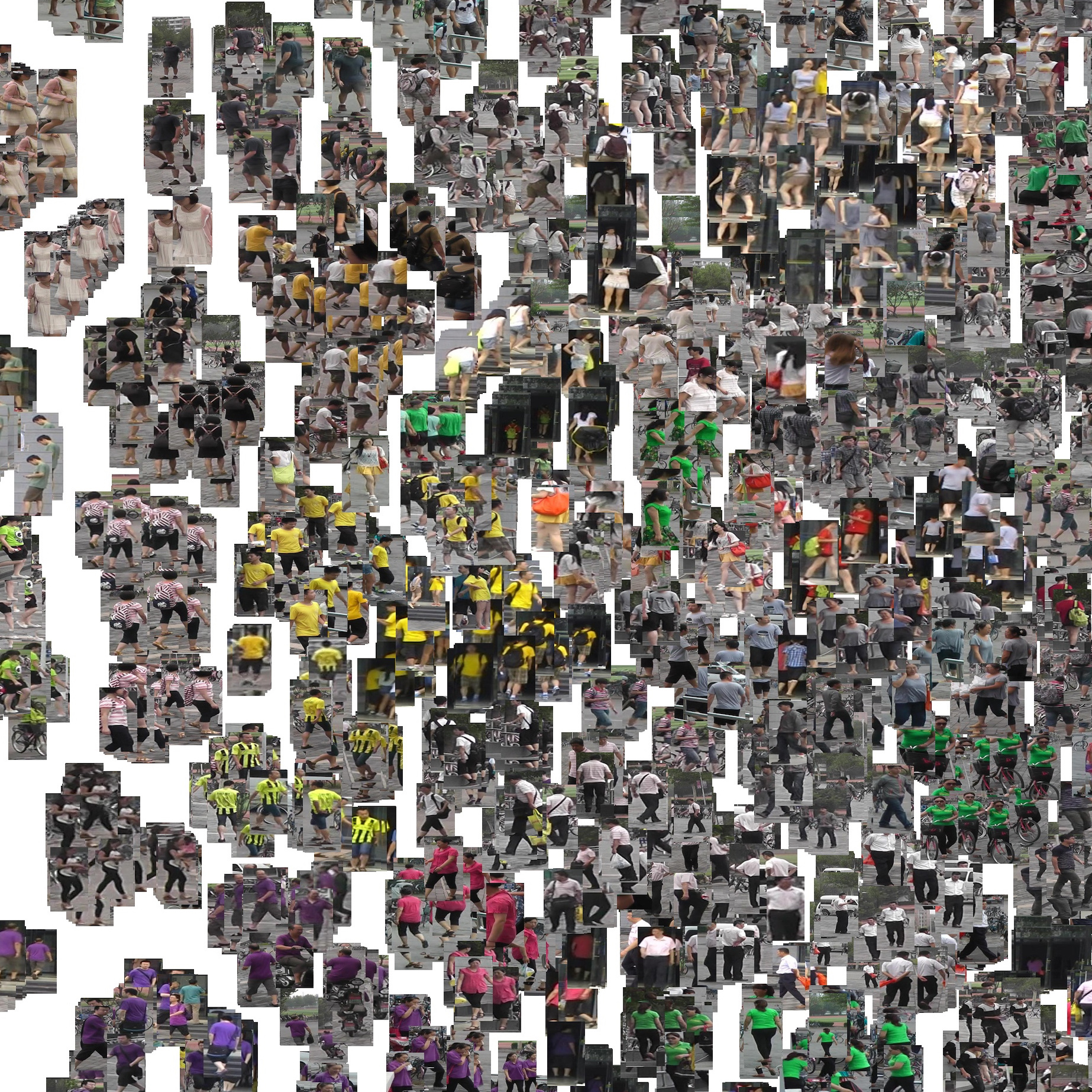}
    \caption[Excerpt of learned embedding visualization on MARS test split]{%
        %This plot shows an
        Excerpt of the learned embedding on the MARS test
        split generated with t-SNE~\cite{van2014accelerating}.
        %Images that are close in the image are also close in representation
        %space.
        %The identities shown here have not been used during training.
    }
\label{fig:mars-embedding}

\figurespacing{}
\end{figure}

\figurename~\ref{fig:market-query-good} and~\ref{fig:market-query-bad}
show a series of exemplary queries computed from the Market
1501 %~\cite{zheng2015scalable}
test gallery.
%together with the five most similar ranked and the five most dissimilar ranked
%images as returned by the cosine softmax network.
The queries shown in \figurename~\ref{fig:market-query-good} represent a
selection of many identities that the network successfully identifies by
nearest neighbor search.
In many cases, the feature representation is robust to varying poses as
well as changing background and image quality.
\figurename~\ref{fig:market-query-bad} shows some challenging queries and
interesting failure cases. For example, in the second row the network seems to
focus on the bright handbag in a low-resolution capture of a woman.
The top five results returned by the network contain four women
with colorful clothing.
In the third row the network fails to correctly identify the gender of the
queried identity.
In the last example, the network successfully re-identifies a person that is
first sitting on a scooter and later walks~(rank 4 and 5), but also
returns a wrong identity with similarly striped sweater~(rank 3).
A visualization of the learned embedding on the MARS test split is shown in
\figurename~\ref{fig:mars-embedding}. % for visual inspection.

\section{Conclusion}

We have presented a re-parametrization of the conventional softmax classifier
that enforces a cosine similarity on the representation space
when trained to identify the individuals in the training set.
Due to this property, the classifier can be stripped of the network after
training and queries for unseen identities can be performed using
nearest-neighbor search.
Thus, the presented approach offers a simple, easily applicable alternative for
metric learning that does not require sophisticated sampling strategies.
%In our experiments, training in this regime consistently improves the
%test performance by a small margin.
In our experiments, training in this regime provided a modest gain in test
performance.
%The experiments have shown training in this regime provides a modest gain in
%test performance.
%Yet, we achieve competitive results on two large-scale person re-identification
%datasets.
While the method itself is general, our evaluation was limited to a very
specific application using a single light-weight CNN architecture.
In future work, the approach should be further validated on more datasets
and application domains. Such an evaluation should also include larger
capacity architectures and pre-training on ImageNet.

{\small
\bibliographystyle{ieee}
\bibliography{paper}
}

\end{document}